%% file: main.tex
\documentclass[onecolumn]{article}
\usepackage[utf8]{inputenc}
\usepackage{xcolor}
\usepackage{hyperref}
\usepackage{graphicx}
\usepackage{subcaption}
\usepackage[misc]{ifsym}
\usepackage{amsmath}
\usepackage{amsfonts}
\usepackage{authblk}
\usepackage{float}
\usepackage{lipsum}
\usepackage{booktabs}
\usepackage{multirow}

\title{Few-shot incremental learning in the context of solar cell quality inspection}
\author[1]{Julen Balzategui}
\author[1]{Luka Eciolaza}
\date{June 2021}

\affil[1]{Mondragon Unibertsitatea}

\begin{document}

\maketitle

\begin{abstract}
    In industry, Deep Neural Networks have shown high defect detection rates surpassing other more traditional manual feature engineering based proposals. This has been achieved mainly through supervised training where a great amount of data is required in order to learn good classification models. However, such amount of data is sometimes hard to obtain in industrial scenarios, as few defective pieces are produced normally. In addition, certain kinds of defects are very rare and usually just appear from time to time, which makes the generation of a proper dataset for training a classification model even harder. Moreover, the lack of available data limits the adaptation of inspection models to new defect types that appear in production as it might require a model retraining in order to incorporate the detects and detect them. In this work, we have explored the technique of weight imprinting in the context of solar cell quality inspection where we have trained a network on three base defect classes, and then we have incorporated new defect classes using few samples. The results have shown that this technique allows the network to extend its knowledge with regard to defect classes with few samples, which can be interesting for industrial practitioners.
\end{abstract}
\input{intro}

\input{method}
\input{evaluation}
\input{experiments}
\input{conclusions}

\bibliographystyle{ieeetr}
\bibliography{biblio.bib}

\end{document}

%% file: intro.tex
\section{Introduction}{

In the last years, image analysis with Deep Learning (DL) methods has shown its full potential regarding precision, robustness, and flexibility in different manufacturing scenarios such as the quality inspection. Among the available approaches, supervised learning has been the one that has reported the most precise results with respect to defect detection rates.

However, supervised training requires a substantial amount of annotated data, which in the quality inspections scenario might be difficult to obtain. It is usually hard to gather enough representative samples for each kind of defect to generate a proper dataset for training, and depending on the task (e.g., segmentation), the annotation of the great amount of data can be an arduous task to accomplish. 

A possible alternative to avoid the need for defective samples may be to approach the problem from an anomaly detection perspective \cite{TSAI2021101272, julen2021anomaly,8248461}, where instead of creating the training set from defective samples, the set is generated using defect-free samples that are more accessible in a production line. However, as the objective of the unsupervised approach is to detect anomalous patterns which encompasses defective and non-defective patterns, it usually reports lower detection rates than supervised models that are strictly trained to detect only defective features. Apart from that, unsupervised anomaly detection approaches mainly focus on separating anomalous samples from defect-free samples rather than identify each anomalous pattern which might be key in an industrial context. Both, a severe defect and a less severe one may fall into the same anomalous category but the piece in which the each of the defect was detected will follow a very different path.

Other solution to avoid the lack of defective samples is the technique called Transfer Learning, which involves reusing the feature extraction capabilities of already trained networks as a way of initializing models for training them in other domains\cite{Hinton2015}. This method allows practitioners \cite{Michieli_2019_ICCV,ferguson2018detection, AKRAM2020175, Zyout} to focus the training of the neural network on learning how to extract domain specific features rather than spending time also in the general features that are shared across domains, which will reduce to a certain level the amount of required training data. However, the success of this technique is directly proportional to the similarity between the source and target domains which in the case of industrial setting can limit its applicability.

Lately, the technique called Few-shot learning has been gaining traction as it allows models to classify instances of never seen classes by just employing few samples of them. The first work in the context of image segmentation was \cite{Shaban2017} which was afterwards followed by other works \cite{Rakelly2018c, Yang2019a, Dong2018a, Zhang2018, Zhang2019a} that proposed variations and improvement over the initial idea. The original work consisted in employing a two-branch based neural network, where a base branch focused on segmenting a new class on a query image while it was guided by parameters coming from the auxiliary branch that were obtained after processing a support set composed of annotated samples of the new class. However, this configuration was not designed taking into account the incremental learning perspective but for a moment specific few-shot case. The network would be guided to segment the new class without paying attention if the model would still be capable of performing well on old classes. From an industrial inspection point of view, a continual adaptation is a more interesting design since new defects, as well as the old ones, should be detected. It might happen that during production a feature that previously was not taken into account for the quality inspection has been revealed to be crucial for the final product's overall quality, thus, the company now wants to incorporate it into the already built inspection model.

In these kind of cases, the few-shot incremental learning will come in handy regarding the update of the inspection model as there are usually few samples available that contain such novel feature or due to the lack of annotations in historical data.

In this work, we experiment with the incremental few-shot learning proposed in \cite{siam2019} in the industrial context of solar cell inspection where we try to update a model that has been trained on three base defect classes (i.e., cracks, micro-cracks, and finger interruptions) and incorporate new defects (i.e., black spots and bad soldering) such that the final model is able to segment five different defects. In this way, we will add another feature into the proposed anomaly detection based methodology \cite{julen2021anomaly} incorporating an adaptation stage for the supervised model to make it able to detect new types of defect classes that can arise during production, or also, incorporate features that were not previously considered important but for now on it should be controlled.

}

The remaining of paper is organized as follows: Section \ref{sec:method} presents how the method works and how we have adapted for our use of case. Section \ref{sec:exp_set} described the dataset, hardware and software specifications, and metrics used in the experiments. Section \ref{sec:experiments} describes the experiments we carried out and the results we obtained. And finally, Section \ref{sec:conclusions} provides some conclusions and possible future works we identified in this work.

%% file: method.tex
\section{Few-shot Incremental Learning}{
\label{sec:method}

The proposal in \cite{siam2019} consists is an adaptation for image segmentation of the "weight imprinting" technique proposed in \cite{Qi2018} for image classification. In \cite{Qi2018}, they establish the connection between the softmax cross-entropy function used in classification problems and proxy-NCA loss function \cite{movshovitz2017no} proposed for metric learning problems.

The proxy-NCA loss function was proposed as a way to reduce computation burden, and thus speed up the time to converge, of methods applied for metric learning using \textit{triplets loss}. Triplets loss aims to minimize the distance between similar points and maximize distance with dissimilar points in a set using pairs of triplet points (anchor point, a positive similar point, and a negative dissimilar point). In order to reduce the amount of possible triplets combinations, they propose to use sets of \textit{proxies} as the representative point for every class. Thus, instead of computing the loss for all possible triplet combinations, the triplets combinations will be reduced to each point, the positive proxy of the point and, the different negative proxies.

Following this idea, \cite{Qi2018} under the assumption that point vectors and proxy vectors are normalized to the same length, they establish a similarity between the proxy-NCA loss and Softmax cross-entropy loss used in neural networks to train the classifier. They argue that if both vectors are normalized to the same length, minimizing the euclidean distance between a point and its proxy is equivalent to maximizing their cosine similarity, thus, the euclidean distance in proxy-NCA loss can be substituted by the cosine similarity. The substitution makes the loss resemble the Softmax cross-entropy. Following this similarity, they prove that normalized embeddings from the neural network can act as weights for a new class in the final Fully Connected layer of the network, so one example will be sufficient to extract embeddings and perform what they called \textit{imprinting}, extending the number of classes in the classifier.

However, unlike the classification, in the context of segmentation the output embeddings have not a vector shape but a three-dimensional shape containing features from different classes. To adapt the \textit{weight imprinting} from classification to segmentation, \cite{siam2019} propose an architecture that incorporates a Normalized Masked Average Pooling (NMAP) layer where the output embeddings are pixel-wise masked for the new class with a given support set that contains samples with instances of the new class and their corresponding labels, and then are averaged and normalized by procedures in Equation \ref{eq:mask} and Equation \ref{eq:norm}. In this way, just relevant features for the new class remain in embeddings that will be used as the proxy for the new class $c$.

\begin{equation}
    P_{c} = \frac{1}{k} \sum_{i=1}^{k} \frac{1}{N}\sum_{x\in X}F^{i}(x)Y^i_{c}(x),
    \label{eq:mask}
\end{equation}

\begin{equation}
    \tilde{P_{c}} = \frac{P_{c}}{\lVert P_{c} \rVert_{2}},
    \label{eq:norm}
\end{equation}
where $Y^{i}_{c}$ is a binary mask for the $i^{th}$ image with the new class $c$ in the support set, $F^{i}$ is the feature maps for $i^{th}$ image. $X$ is the set of all possible spatial locations and $N$ is the number of pixels that are labeled as foreground for class $c$. Moreover, they extract features at different layers in the network when computing the proxies so information at several resolutions is taken into account, and thus improve the final segmentation result. The network architecture is shown in Figure \ref{fig:archi_ori}. The base architecture is VGG-16 \cite{simonyan2014very}, with additional skip connections to extract different resolution embeddings which are composed of 1x1 convolution layers. The embeddings at these layers are mapped to the label space to then perform the weight imprinting for the new class.

\begin{figure*}[!ht]
     \centering
     \includegraphics[width=0.8\textwidth]{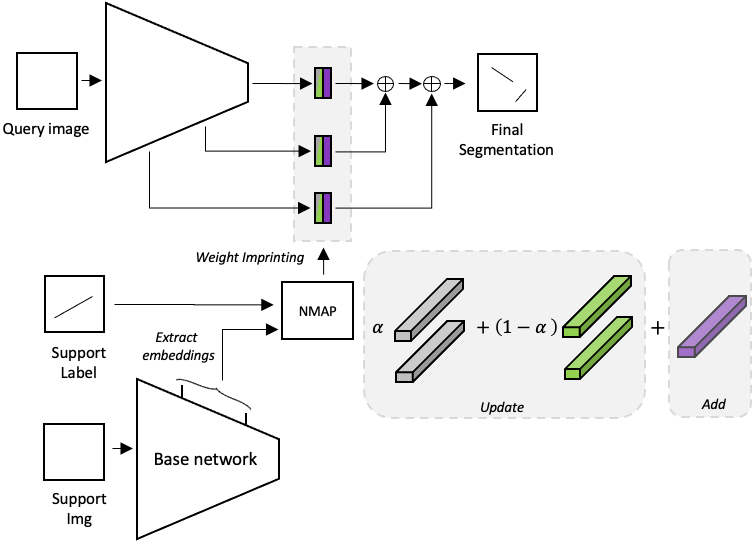}
     \caption{FCN based architecture for the weight imprinting}
     \label{fig:archi_ori}
\end{figure*}

In addition to incorporating weights of the new classes, they consider the continual learning scenario by updating weights from the learned classes with the information coming from the support set at every imprinting procedure. When an instance of a learned class is present in the support set along with the new classes, the embeddings from that instance are also extracted. Then, at imprinting, apart from incorporating the proxies of the new classes, the weights from old classes are updated with their correspondent proxies by Equation \ref{eq:adaptation}.

\begin{equation}
    \tilde{W}_{c} = \alpha \tilde{P_{c}} + (1 - \alpha)W_{c}
    \label{eq:adaptation}
\end{equation}
where $\tilde{P_{c}}$ is the normalized proxy for the class $c$, $W_{c}$ are the weights from the previously learned classes at the 1x1 convolution layers, and $\alpha$ it the updating rate. In this way, as new samples from the last imprinted classes arise, the features that were not present when the first imprinting was performed can be incorporated into the network. Thus, the weight related to the new class can be consolidated.

In this work, we have employed the architecture in \cite{siam2019} in our industrial dataset, and also experiment with other architecture illustrated in Figure \ref{fig:unet_arch} which is based on U-net network proposed in \cite{DBLP:journals/corr/RonnebergerFB15} with the objective of improving the results. After some first tests, we saw that the original architecture yielded poor and coarse segmentation results regarding the shape of big defects and number of detected defects. Precise results are key to establish the severity of the defect and determine if the cell need to be completely discarded or could be repaired to put it back in the panel.

The original U-Net architecture was designed to follow an encoder-decoder shape with skip connections between the blocks in the encoder and decoder parts. These connections were implemented to allow the network to use the full potential of the features for the segmentation. For this paper, the network was extended by adding more skip connections composed of 1x1 convolutional layers to perform the weight imprinting. Figure \ref{fig:unet_arch} shows this extension where the original connections are in gray and the new ones in green and purple.

\begin{figure*}[!ht]
     \centering
     \includegraphics[width=0.8\textwidth]{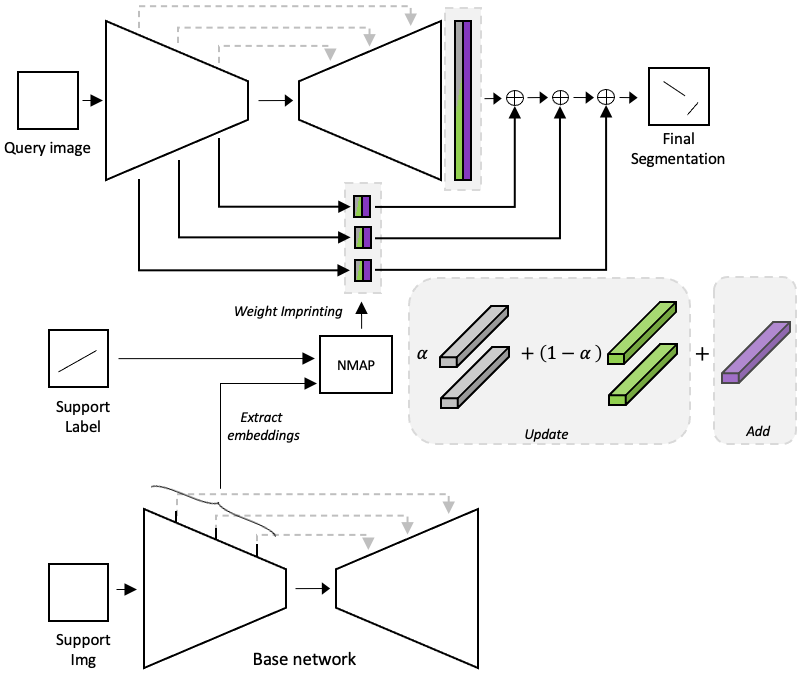}
     \caption{U-net based architecture for the weight imprinting}
     \label{fig:unet_arch}
\end{figure*}

}

%% file: evaluation.tex
\section{Experimental setup}{
\label{sec:exp_set}
In this section, how the experiments have been designed by means of the network configurations, the dataset employed for evaluation, and hardware and software specifications used in the experiments will be described. Although the results have been mainly evaluated at qualitative level a quantitative evaluation has also been done. Because of this, the metrics employed in the quantitative analysis are going to be described.

\subsection{Dataset}{
The dataset used in an industrial dataset provided by Mondragon Assembly S.Coop. that is composed of Electroluminescence images of monocrystalline solar cells obtained from a real industrial production line. Electroluminescence is a widely used technique during the industrial quality inspection of solar panels. The technique consists in putting the panels under electrical current and capturing high resolution images of the light that cells emit caused by the phenomenon of the Electroluminescence. As just the areas that were not supposed to conduct the electricity should appear dark in the images, those areas that have turned non-conductive because of the defects can be spotted on \cite{Fuyuki2009}. The cells that compose the dataset are monocrystalline cells that contain different defects: cracks, microcracks, finger interruptions, black spots and bad soldering. Some examples of these defects are illustrated in Figure \ref{fig:dataset_samples}.

\begin{figure*}[!ht]
    \centering
    \includegraphics[width=\textwidth]{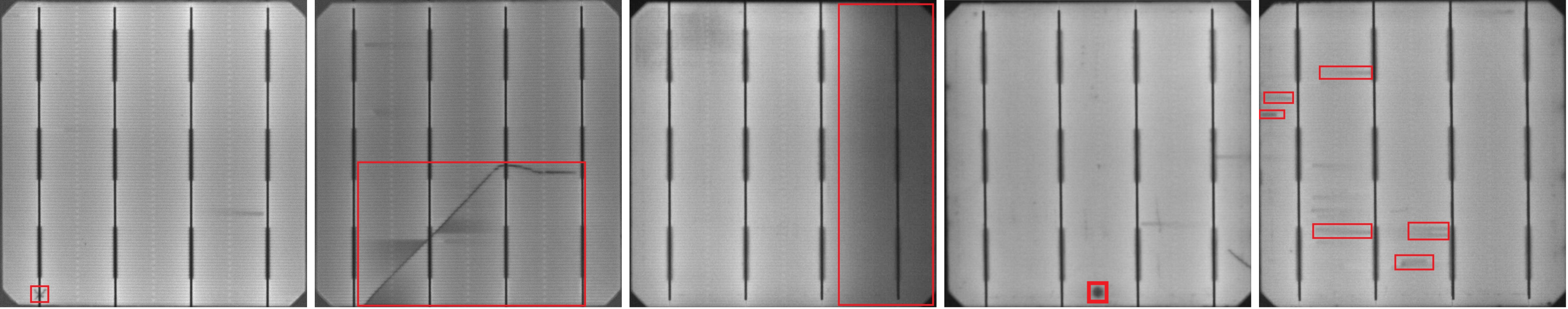}
    \caption{Different type of defects highlighted in Electroluminescence images of solar cells. The defects are: a) microcrack, b) crack, c) bad soldering, d) black spots and e) finger interruptions.}
    \label{fig:dataset_samples}
\end{figure*}

Each sample $x$ in the dataset has its pixel-wise annotation sample $y$ that includes semantic labels for the different defects $C$. The samples in the dataset have been used for different purposes: train the base network, perform the weight imprinting and evaluation. Thus three different sets were generated. 1) the training set $D_{train}$ was composed of samples that contained cracks, finger interruption, and microcracks defect classes (i.e., base classes $c_{b}$ and was used to train the base network. 2) The support set $D_{supp}$ that was composed of samples with black spots and bad soldering (i.e., new classes $c_{n}$). Some of the samples in the training set also contained instances of black spots, so in order to consider black spots as a new class for imprinting, the instances of black spots in the training sample labels were removed, and thus, considered as background. 
% \TODO{$D_{train} \cap D_{supp_{black\_spot}} $. NO SE COMO PONER ESTO}. 
In the support set instead, instances of the base classes were not removed to simulate the continuous learning scenario. 3) Regarding the test set, it contained samples that were put aside just for evaluation that considered all types of defects. This latter set served to evaluate how the imprinting affected the network in the segmentation of the new classes, and also if could impact negatively the capability of detecting old classes. The test set also had defect-free samples to evaluate the weight imprinting in these samples. Table \ref{tab:dataset} shows the sample distribution for all the classes that include the mentioned defects and also defect-free samples.

\begin{table}[!ht]
\def\arraystretch{1.2}
\centering
\begin{tabular}{lcccccc}
\toprule

\multicolumn{1}{c}{\textbf{}} &  &\textbf{Train}&\textbf{Imprin.}&\textbf{Test}&\textbf{Total}\\ \hline
\multicolumn{2}{l}{\textbf{Defect-free}} &- &-&375&375\\ 
\multicolumn{2}{l}{\textbf{Defective}} &  &&&386\\
\multicolumn{2}{r}{Crack} &14&-&4&18 \\ 
\multicolumn{2}{r}{Microcrack} &192&-&48&240 \\ 
\multicolumn{2}{r}{Finger inter.} &93&-&24&117 \\ \bottomrule
\multicolumn{2}{r}{Black spots*} & -&4&3&7 \\
\multicolumn{2}{r}{Bad Soldering} &-&2&2&4 \\
\end{tabular}

\caption{Sample distribution in the dataset. 
}
\label{tab:dataset}
\end{table}

}

\subsection{Metrics}{
As stated at the beginning of the section, the results have been mainly evaluated at qualitative level. Nonetheless, how the imprinting operation might affect network image-level defect detection capability has been also evaluated. For this evaluation the metrics Precision, Recall, and Specificity, defined in Equations \ref{eq:precision}, \ref{eq:recall}, and \ref{eq:specificity} were used.

\begin{equation}
    Precision = \frac{TP}{TP + FP}
    \label{eq:precision}
\end{equation}
\begin{equation}
    Recall = \frac{TP}{TP + FN}
    \label{eq:recall}
\end{equation}
\begin{equation}
    Specificity = \frac{TN}{TN + FP}
    \label{eq:specificity}
\end{equation}
where TP stands for True Positive, TN for True Negative, FN for False Negative, and FP for False Positive. In this work, defective samples belong to the Positive class and defect-free samples to the Negative class.

In order to compute the metrics, the whole prediction must be classified as defective or defect-free. As the smallest defect instances are about 40 pixels in size, it was decided to classify the cells as defective if the prediction contained more than 20 defective pixels.

}

\subsection{Hardware and software}{
All the experiments were performed using the publicly available code in github \footnote{https://github.com/MSiam/AdaptiveMaskedProxies}. Regarding the hardware, the training of the networks were performed on a server with one Nvidia GeForce RTX 2080 GPU and CUDA 10.}

}

%% file: experiments.tex
\section{Experiments}{
\label{sec:experiments}

In order to check the applicability of architecture presented in \cite{siam2019} in our dataset, an initial training was performed where the network was trained on the base classes in the training set, and then executed on the test samples that contain cracks, finger interruptions and microcracks. The results of this first experiment is shown in Figure \ref{fig:results_imprinting_fcn}. 

\begin{figure*}[!h]
    \centering
    \includegraphics[width=0.70\textwidth]{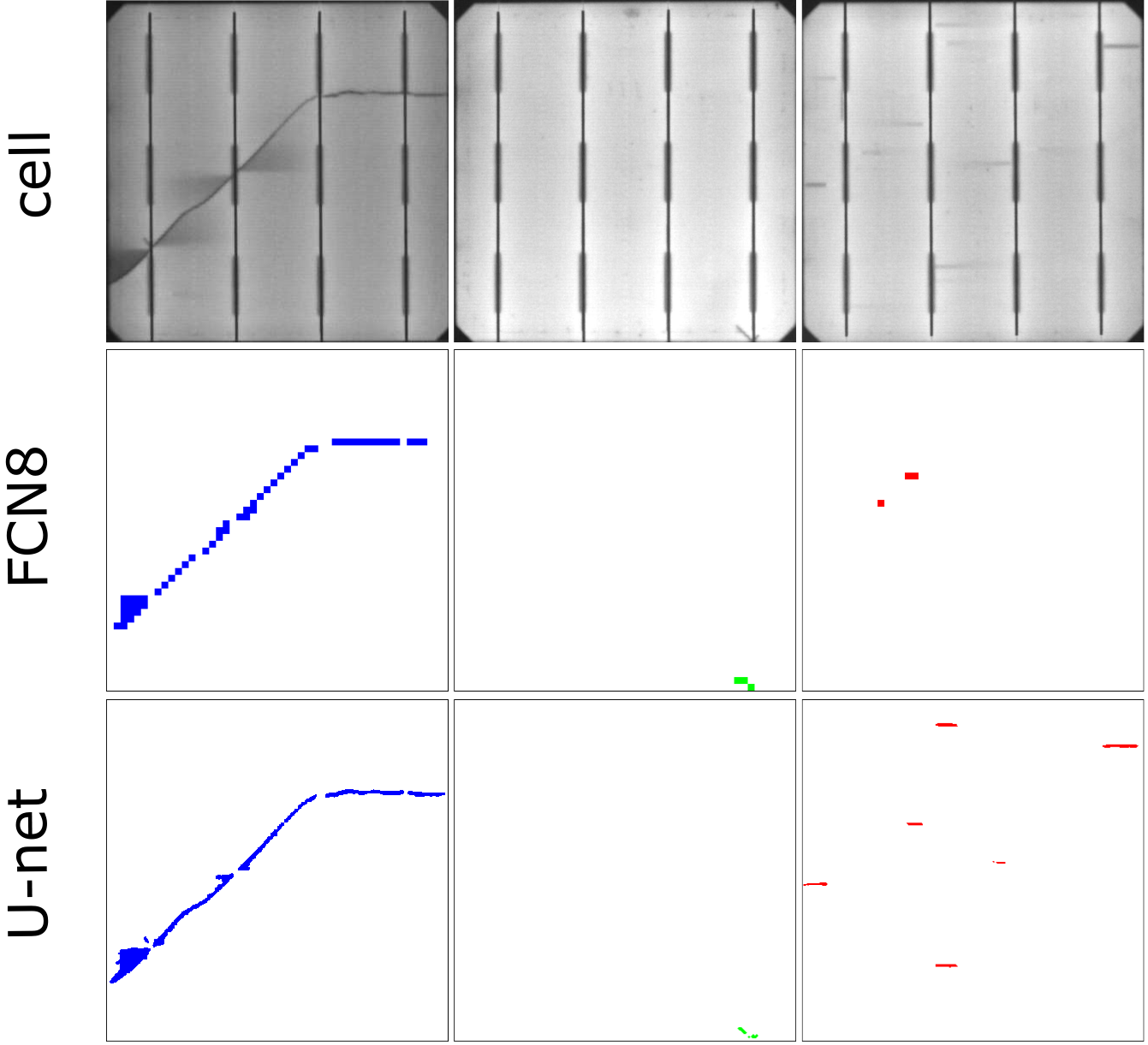}
    \caption{Results on the base defect classes (cracks, microcracks and finger interruptions) with the original network and with the U-net based architecture}
    \label{fig:results_imprinting_fcn}
\end{figure*}

These first results show that the original network was able to locate almost all defects in the cells. In the case of small and light defects, such as the finger interruptions in the third sample in the figure, the network could not segment them. In addition to these, even though bigger defects were detected, the network output a coarse segmentation. After these results, the architecture was substituted with U-net based network, and the same training and testing were performed.

As mentioned above, U-net was extended with extra skip connections to then perform the weight imprinting. The rest of the architecture with respect to the blocks at the encoder and decoder parts was kept as in \cite{DBLP:journals/corr/RonnebergerFB15}. The training was performed using the cross-entropy loss and RMSprop optimization function. In this case, weighted cross-entropy was required for training to alleviate the effect of the unbalance between defective and non-defective pixels, especially for the cases of microcrack and finger interruptions where the defective pixels suppose less than 0.1\% of the pixels in the samples. As it can be seen in Figure \ref{fig:results_imprinting_fcn}, the segmentation results with U-net were more refined than with the original network, and also, the lack of detection with respect to the smaller defects was solved.

After the base training, two sequential imprinting operations were performed: first the black spots defect class was incorporated, and then, the bad soldering defect class was incorporated. Figure \ref{fig:new_classes_example_imprinting} exhibits two examples that contained instances of these two defect classes.

\begin{figure}[!h]
    \centering
    \includegraphics[width=0.7\columnwidth]{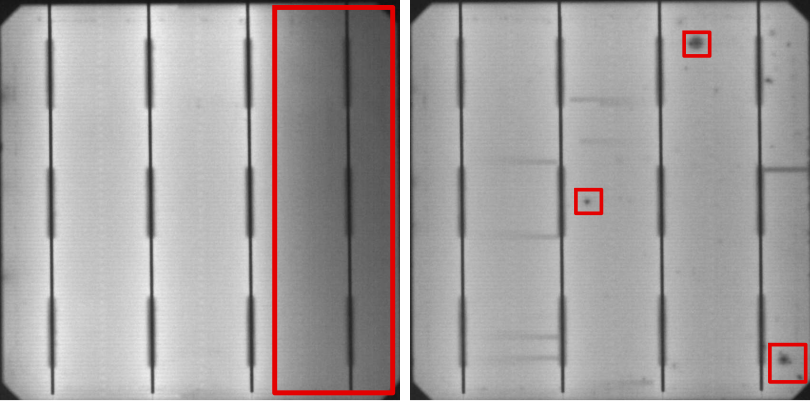}
    \caption{Bad-soldering and black spots defect classes examples highlighted with bounding boxes.}
    \label{fig:new_classes_example_imprinting}
\end{figure}

It the end, we obtained a network that can segment 5 classes (3 base + 2 new). For this operation, the samples allocated for imprinting in Table \ref{tab:dataset} were employed. In order to check the impact of every imprinting, after every imprinting operation, the network was executed on the entire test set. Some samples from these executions are shown in Figures \ref{fig:results_imprinting_unet_base} and \ref{fig:results_imprinting_unet_new}. The samples in Figure \ref{fig:results_imprinting_unet_base} illustrate samples with the base classes (cracks, microcracks, and finger interruptions), while the samples in Figure \ref{fig:results_imprinting_unet_new} show some instances of the new classes (i.e. bad soldering on the left cell and black spots on the right cell).

\begin{figure*}[!h]
    \centering
    \includegraphics[width=0.75\textwidth]{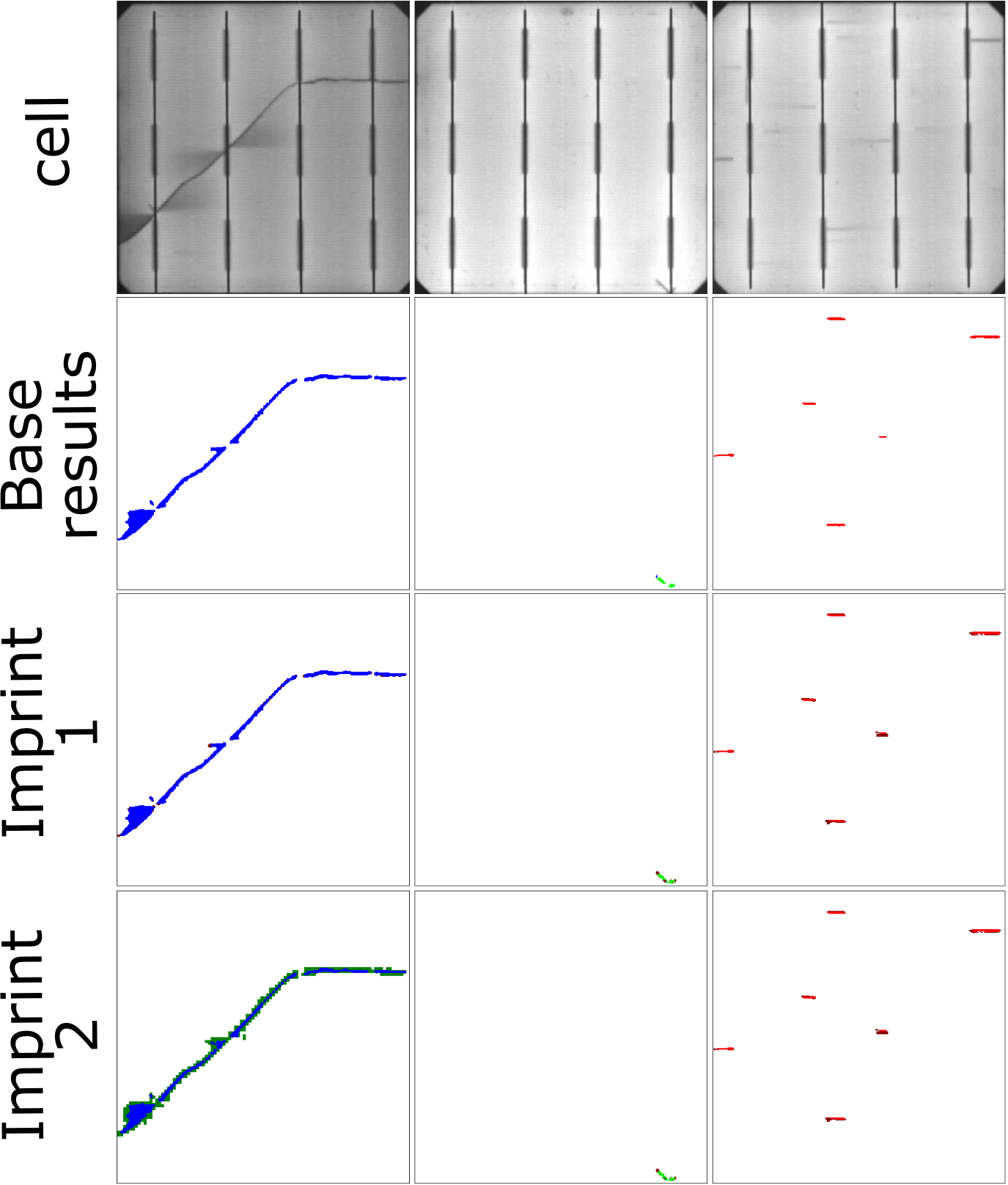}
    \caption{Segmentation results on samples with base defect classes before imprinting and after each imprinting. The colors of the classes are: blue-cracks, light green-microcracks, red-finger interruptions}
    \label{fig:results_imprinting_unet_base}
\end{figure*}

\begin{figure*}[!h]
    \centering
    \includegraphics[width=0.85\textwidth]{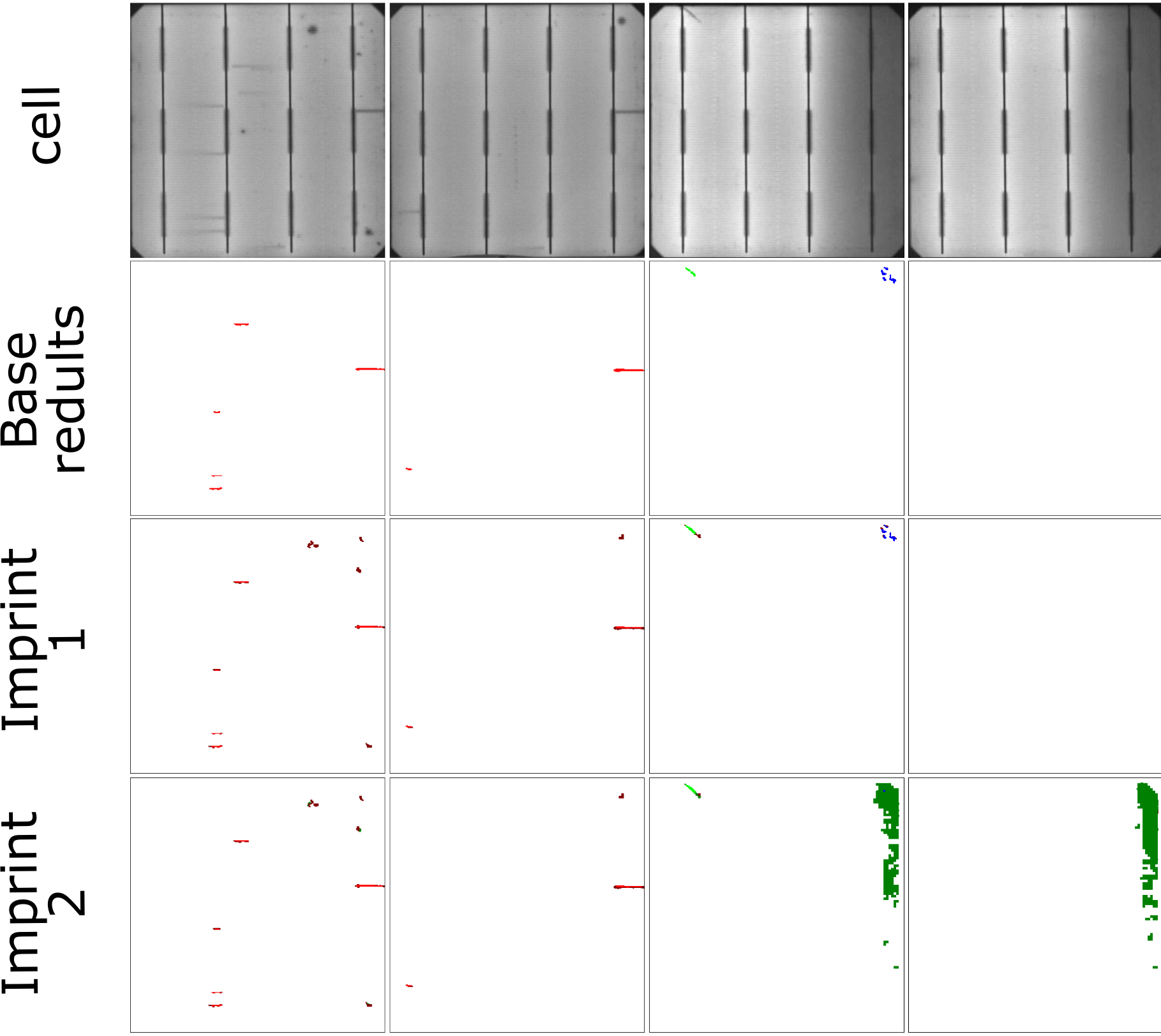}
    \caption{Segmentation results on samples with new defect classes before imprinting and after each imprinting. The colors of the classes are: brown-black spots, and dark green-bad soldering.}
    \label{fig:results_imprinting_unet_new}
\end{figure*}

Figure \ref{fig:results_imprinting_unet_base} shows that the base network segmented the base classes with high precision. As expected, the segmentation was more accurate when the defects were bigger and present higher contrast with respect to the cell background. This is especially visible in the finger interruptions in the third sample, where the darker instances are more thoroughly segmented than the lighter ones. With respect to the samples with the new classes in Figure \ref{fig:results_imprinting_unet_new}, the black spots instances were not segmented in any of the samples. In the case of the bad soldering class instead, the network very vaguely detected some defective features in the top-right corner of the cell that belong to a bad soldering defect.

After the first imprinting (i.e. black spots class addition), the instances from black spots were segmented as it can be appreciated in the first and second column samples in Figure \ref{fig:results_imprinting_unet_new}. In this case also, the darker the defect, the better the segmentation was. However, the imprinting affected the segmentation of the base classes as can be seen in samples from the second row in Figure \ref{fig:results_imprinting_unet_base}. The network segmented some pixels at the borders of the "older" defects as they belonged to pixels from a black spot class, for example, the microcrack in the second sample. 

After the second imprinting with the bad soldering class, the same effect as the first imprinting was appreciated. The network was able to segment the bad soldering instances as it can be observed in the last two samples in Figure \ref{fig:results_imprinting_unet_new}, but at the same time it implied the appearance of some false positives around the previous segmentation results. Nonetheless, as it is clearly visible, the segmentation of the bad soldering was poorly accomplished.

Overall, the weight imprinting allowed the network to segment new classes using just a few defective samples from each new classes (2 for bad soldering and 4 for black spots). Nevertheless, after the imprinting, the network started mistakenly classifying certain areas around the "older" defects as they were new defect class instances.

In addition to the qualitative results, quantitative metrics were also computed to also illustrate the effect of the imprinting in the model performance. The results for these metrics are in Tables \ref{tab:results_image_few} and \ref{tab:results_per_defect_instances}. Table \ref{tab:results_image_few} represents the metrics at image level as the problem was a binary cell classification, defective (all defects) vs defect-free. And Table \ref{tab:results_per_defect_instances} shows the detection rate per defect class, i.e. from all the instances of each class in the cells what percentage was detected. As with the segmentation results, the results in the tables represent the performance with the the base network and after each of the imprinting operation.

\begin{table}[!ht]
\centering
{\renewcommand{\arraystretch}{1.}
\begin{tabular}{lccc}

\toprule
\textbf{}      &    \textbf{Recall} & \textbf{Precision} & \textbf{Specificity}\\ \midrule
\multicolumn{1}{l}{\textbf{Base network}} & 88 & 86 & 99\\
\multicolumn{1}{l}{\textbf{1st Imprint.}} & 96 & 81 & 96\\
\multicolumn{1}{l}{\textbf{2nd Imprint.}} &  97 & 77 & 94\\\bottomrule

\end{tabular}}
\caption{Results at image level before and after each imprinting.}
\label{tab:results_image_few}
\end{table}

\begin{table}[!ht]
\def\arraystretch{1.2}
\centering
\begin{tabular}{l|c|ccc}
\toprule
              & \begin{tabular}[c]{@{}c@{}}\textbf{Orig. model}\\ \textbf{(FCN8)}\end{tabular} & \begin{tabular}[c]{@{}c@{}}\textbf{Base Model}\\ \textbf{(Unet)}\end{tabular} & \multicolumn{1}{l}{\textbf{Imprint. 1}} & \multicolumn{1}{l}{\textbf{Imprint. 2}} \\ \hline
\textbf{Crack}         & 100\%                                                        & 100\%                                                       & 100\%                          & 100\%                          \\
\textbf{Microcrack}    & 64\%                                                         & 71\%                                                        & 90\%                           & 90\%                           \\
\textbf{Finger inter.} & 65\%                                                         & 88\%                                                        & 90\%                           & 90\%                           \\
\textbf{Black spots}   & 0\%                                                          & 0\%                                                         & 77\%                           & 77\%                           \\ 
\textbf{Bad Solder.}   & 0\%                                                          & 0\%                                                         & 0\%                            & 100\%                          \\
\bottomrule
\end{tabular}

\caption{Percentages of detection of each model per defect class. Note that the imprinting columns results refer to the ones from U-net base network.}
\label{tab:results_per_defect_instances}
\end{table}

As it can be observed, in general terms the imprinting allowed the network to detect more defects within the cells, and thus, detect more defective cells making the Recall increase after each imprinting operation. Also, the imprinting made the network segment some instances of the base defects that were initially very vaguely segmented, turning some False Negative into True Positive cases. If the results are broken down by defect classes as shown in Table \ref{tab:results_per_defect_instances}, the effect of the imprinting is easily appreciable in the case of the new classes, but also in the case of samples with microcracks. In the case of the latter, the models went from detecting about a 71\% of the microcrack instances to around 90\% of the samples and from 88\% to 90\% finger interruptions instances.

The improvement in microcracks is given by those instances that are very small and in some case resemble black spots. In these cases, previous to the imprinting, the network classified the pixels corresponding to the defect as background, thus they were considered as undetected. But after the imprinting, these pixels were classified as black spots and microcracks. From the defect detection point view, as the main objective is to highlight the presence of defects in the cells, the defect instances were considered detected as reflected in the results. However, from the pixel segmentation point of view, as the mentioned when describing Figure \ref{fig:results_imprinting_unet_new} the pixel classification got a bit noisy after the imprinting operations.

In addition, the network started to segment certain areas in defect-free samples as they were defective (mainly dark areas that resemble black spots). After a manual analysis of these cases, it was found that some of them could be classified as defective as they really contained these black spots, in others instead, they were just False Positive cases. In total from the 375 defect-free samples, there were 15 samples that could be considered as defective based on the presence of black spots. Taking this into account, these 15 cases were put aside and the remaining 360 defect-free samples were employed for the metrics computation. As mentioned, the imprinting operation makes model start to segment certain areas within the defect-free samples as they were defective areas. This translates to more False Positive cases making the Specificity as well as Precision to decrease as reflected in Table \ref{tab:results_image_few} and Table \ref{tab:results_per_defect_instances}.

}

%% file: conclusions.tex
\section{Conclusions}{
\label{sec:conclusions}
In this paper, we have experimented with the weight imprinting technique in the context of solar cell manufacturing that allows the neural network to increase the classes that it is capable to segment just with few defective samples. We have explored changing the architecture from the original proposal which has resulted in more precise segmentation. This technique adds another step in the methodology presented in \cite{julen2021anomaly} which completes the process to develop an inspection system by first using defect-free samples to generate an anomaly inspection model, then which will serve as automatic labeling to train a more accurate model, and now update the final model by incorporating new defect classes with few samples.

The experiments in this work have focused on sequentially incorporating two additional defect classes in the network on top of the initial three base classes. The results have shown that the weight imprinting technique allows the network to expand its capabilities in defect detection. However, after the imprinting is performed, the segmentation with respect to the base classes is affected as it starts segmenting areas around base defect instances as they belong to new defect areas. This effect is appreciable when calculating model performance metrics before and after each imprinting. Nevertheless, the option to increase the number of classes that a network can detect with just a few samples is an interesting feature for industrial cases as it is commonly difficult to have access to such defective samples to perform a proper supervised training.

The results show that this method has the potential to overcome the lack of defective samples that industrial scenario might face and allows the network to increment its segmentation capabilities to respect to new defect classes. However as future working lines, it would be necessary to alleviate the effect of the imprinting in the network.
}